\documentclass[letterpaper,10 pt,conference]{ieeeconf}  
\IEEEoverridecommandlockouts                              
\usepackage{graphics}
\usepackage{textcomp}
\usepackage{amsmath} % assumes amsmath package installed
\usepackage{mathtools}% Loads amsmath
\usepackage{amssymb}  % assumes amsmath package installed
\usepackage{url}
\usepackage{commath}
\usepackage{caption}
\usepackage{array} 
\usepackage{color} 

\usepackage{footnote}
\makesavenoteenv{tabular}
\makesavenoteenv{table}
\makesavenoteenv{figure}

\usepackage{siunitx}
\usepackage{subcaption}
\usepackage{tikz}
\usepackage{booktabs}
\usepackage{multirow}
\usepackage{graphicx}
\usepackage{adjustbox}
\usepackage{comment}
\usepackage{dblfloatfix}

\usepackage{color}
\usepackage{colortbl}
\usepackage{acronym}
\acrodef{HRI}[HRI]{human-robot interaction}
\acrodef{ROS2}[ROS 2]{\textit{Robot Operating System 2}}
\acrodef{ROS}[ROS]{\textit{Robot Operating System}}
\acrodef{PPG}[PPG]{Photoplethysmography}
\acrodef{BVP}[BVP]{Blood Volume Pulse}
\acrodef{IBI}[IBI]{Inter-Beat-Interval}
\acrodef{EDA}[EDA]{Electrodermal Activity}
\acrodef{HMRI}[HMRI]{Human Multi-Robot Interaction}
\acrodef{LSL}[LSL]{Lab-Streaming Layer}
\acrodef{MRS}[MRS]{Multi-Robot System}
\acrodef{HR}[HR]{Heart Rate}
\acrodef{EEG}[EEG]{Electroencephalography}
\acrodef{GSR}[GSR]{Galvanic Skin Response}
\acrodef{ST}[ST]{Skin Temperature}
\acrodef{RR}[RR]{Respiratory Rate}
\acrodef{CSV}[CSV]{Comma-Separated Values}
\acrodef{DDS}[DDS]{Data Distribution Service}
\acrodef{ML}[ML]{Machine Learning}

\usepackage{hyperref}

\title{\LARGE \bf Toward a Wearable Biosensor Ecosystem on ROS 2 for Real-time Human-Robot Interaction Systems}
\author{Wonse Jo$^{1}$, Robert Wilson$^{2}$, Jaeeun Kim$^{1}$, Steve McGuire$^{2}$, and Byung-Cheol Min$^{1}$% <-this % stops a space
\thanks{$^{*}$This paper was accepted to the IROS 2021: Workshop on Cognitive and Social Aspects of Human Multi-Robot Interaction (HMRS 2021). The proposed ROS2 package is available to download from \url{https://github.com/SMARTlab-Purdue/ros2-foxy-wearable-biosensors}}
\thanks{$^{1}$Wonse Jo, Jaeeun Kim, and Byung-Cheol Min are with SMART Lab, Department of Computer and Information Technology, Purdue University, West Lafayette, IN 47907, USA \tt\small{jow@purdue.edu, kim2592@purdue.edu, minb@purdue.edu}.}
\thanks{$^{2}$Robert Wilson and Steve McGuire, HARE Lab, Department of Electrical and Computer Engineering, University of California, Santa Cruz, Santa Cruz, CA 95060, USA \tt\small{robert.wilson@ucsc.edu, steve.mcguire@ucsc.edu}.}
}

\begin{document}
\maketitle
%\thispagestyle{empty}
%\pagestyle{empty}
%abstracts -150words
%need to submit with a cover letter 
\begin{abstract}
Wearable biosensors can enable continuous human data capture, facilitating development of real-world \ac{HRI} systems. However, a lack of standardized libraries and implementations adds extraneous complexity to \ac{HRI} system designs, and precludes collaboration across disciplines and institutions. Here, we introduce a novel wearable biosensor package for the \ac{ROS2} system. The \ac{ROS2} officially supports real-time computing and multi-robot systems, and thus provides easy-to-use and reliable streaming data from multiple nodes.
The package standardizes biosensor \ac{HRI} integration, lowers the technical barrier of entry, and expands the biosensor ecosystem into the robotics field. Each biosensor package node follows a generalized node and topic structure concentrated on ease of use. Current package capabilities, listed by biosensor, highlight package standardization. Collected example data demonstrate a full integration of each biosensor into \ac{ROS2}. We expect that standardization of this biosensors package for \ac{ROS2} will greatly simplify use and cross-collaboration across many disciplines. The wearable biosensor package is made publicly available on GitHub at \url{https://github.com/SMARTlab-Purdue/ros2-foxy-wearable-biosensors}.
\end{abstract}

%\begin{keywords}Multi-robot systems; Cognitive human-robot interaction; Deictic cues; Information retrieval; Delayed free recall task \end{keywords}

%%%%%%%%%%%%%%%%%%%%%%%%%%%%%%%%%%%%%%%%%%%%%%%%%%%%%%%%%%%%%%%%%%%%%%%%%%%%%%%%
\section{Introduction}
\label{sec:introduction}
%%%%%%%%%%%%%%%%%%%% v03
%workshop's objective is to explore the cognitive and social aspects in human multi-robot system (MRS) interaction
Along with the advances in robotics technology and emerging various robot platforms, \ac{HMRI} systems have been broadly adopted in various real-world applications \cite{rosenfeld2017intelligent, rizk2019cooperative}. The role of a human (e.g., a human operator) in the \ac{HMRI} system is mainly to physically and remotely deal with unexpected situations/errors and complicated tasks that the robot systems cannot handle due to a lack of experience and hardware specifications, respectively \cite{rodriguez2021human}. However, the existence of humans is not always helpful to improve the overall performance of the \ac{HMRI} system because of the added complexities introduced by widely variable humans 
%could be another variable to increase the complexity of the system
\cite{humann2019human}. Furthermore, the work performance of the human operator can be indirectly and directly related to the operator's affective state (such as emotional and cognitive states) that can be influenced by the pressure of work or personal circumstances. Finally, the human operator can be easily affected by various factors in the \ac{HMRI} system, such as the number of robots, working duration, and operator's trust in automation \cite{parasuraman1997humans}.

Several contemporary methods exist to estimate a human's affective states. One of the common ways is to utilize a vision (i.e., camera) sensor to recognize a person's facial expression \cite{bera2019modelling}, specific gestures \cite{chen2013multimodal}, or analyze the human's emotional states through a gait \cite{narayanan2020proxemo}. 
%Don't start sentences with conjunctions
%But, since humans can easily deceive the system, the camera-based prediction systems are more likely to produce incorrect results. 
One potential source of error in camera-based prediction systems, aside from common problems with machine vision systems in general relating to image quality (lighting, exposure, etc), is the reliance on facial expressions that can be voluntarily controlled. 
%Another way is to utilize biosensors 
Instead of observing biological systems under voluntary control, we propose estimating people's cognitive states from biosignal data. Along with the advances of the smart device and sensor technology, various consumer-available biosensors have been released that have multiple biosensors capable of reading biosignal information, so users can utilize these devices in daily life as auxiliary medical equipment \cite{choi2016smartwatch} or as a fitness assistant \cite{guo2017fitcoach}.
%(e.g., Apple Watch or Samsung Galaxy Watch) have been released to satisfy modern people's needs. For example, the latest smartwatches (such as Apple Watch or Samsung Galaxy watch) 
%have multiple biosensors capable of reading biosignal information, so users can utilize these devices in daily life as auxiliary medical equipment \cite{choi2016smartwatch} or as a fitness assistant \cite{guo2017fitcoach}. 
To learn how to leverage biosignals, numerous studies on how to predict affective states through the wearable biosensor-sourced data are under active development in affective computing fields \cite{yousefi2013motion, Zamkah_2020, leite2013hri}.

% This image will be updated by Wonse Jo
\begin{figure}[t] 
  \centering
  \includegraphics[width=1\columnwidth]{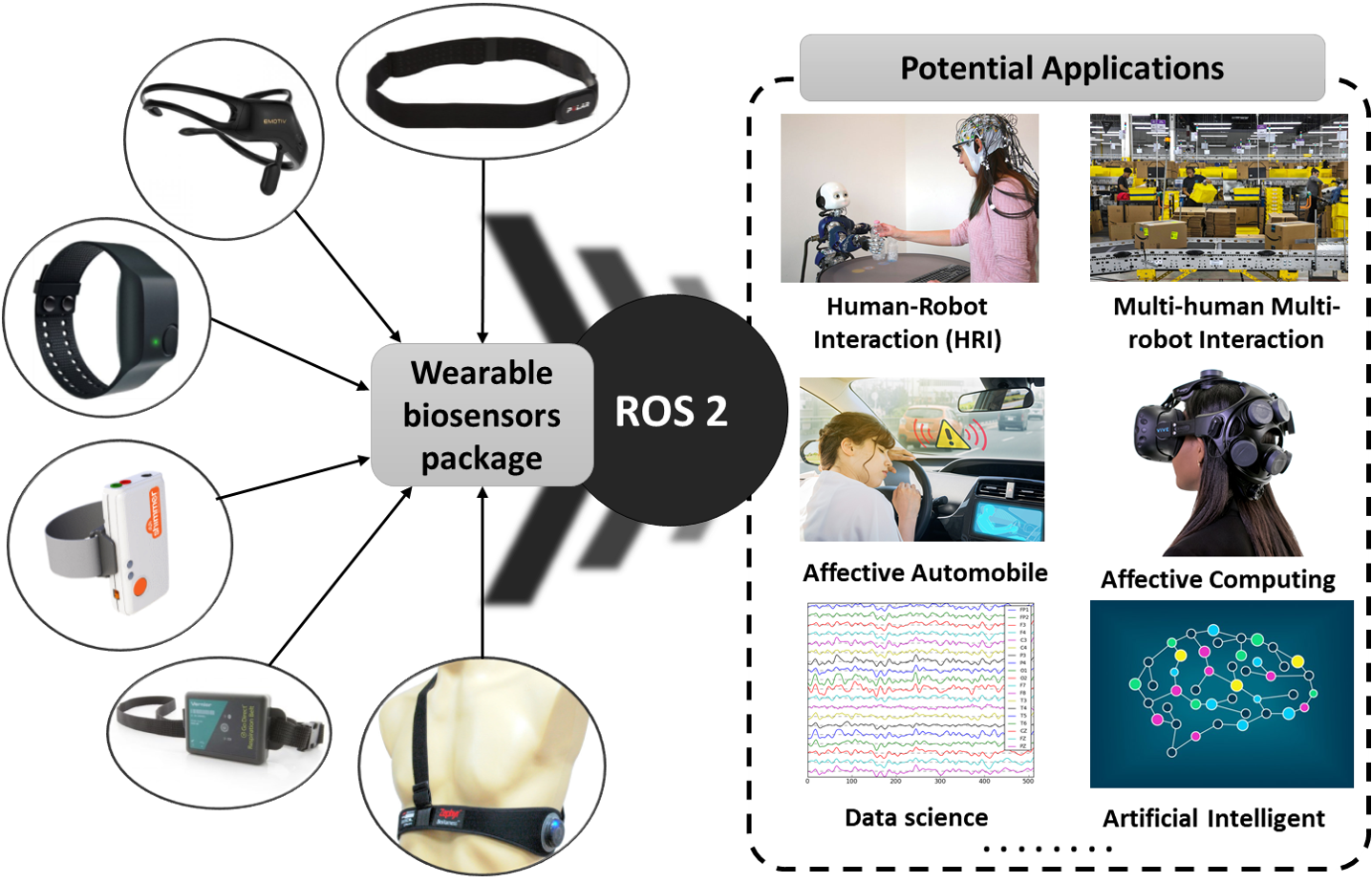}
   \caption{Examples of the potential applications using the proposed wearable biosensor package.}
   \label{fig:title_image}
\end{figure}
In contrast, \ac{HRI} in the robotics field has still barriers to easily use wearable biosensors as part of a robotic system. 
%The biosensors for HRI are required to have \ac{ROS} to merge with robotic systems because the \ac{ROS} is dominated for connecting various robotic hardware or software systems to the main processor. 
In many robotic systems, \ac{ROS} is used as a connectivity layer to manage interactions between many types of hardware and software. Ideally, biosensor data for \ac{HRI} can be integrated using \ac{ROS} as a common framework.
However, most biosensors do not officially support \ac{ROS} since each vendor provides vendor-specific binary packages to read and visualize sensor data that limit widespread adoption within the \ac{HRI} research community. Many researchers who want to use the biosensors with \ac{ROS}, therefore, require additional integration efforts.
%have to take a lot of efforts and time to build a hardware environment and connect them with \ac{ROS}. 

%Real-time has a specific meaning to computer science people - I would recommend use of the term 'online' instead

In this paper, we introduce a new wearable biosensors package for the \ac{ROS2} system. The \ac{ROS2} supports real-time computing, multi-robot systems and various operating systems (e.g., Windows), and provides easy-to-use, reliable and real-time data streaming from multiple nodes. Thus, the integration of the \ac{ROS2} and the biosensors can expand the role of biosensors in the robotics field. The current version of the package (v0.0.1) supports six wearable biosensors that can be used in real-world \ac{HRI} research projects without behavioral constraints caused by limited hardware specifications (e.g., wired devices). The supported wearable biosensors are connected to the \ac{ROS2} system, so that researchers who need to use the wearable biosensors in the \ac{HRI} field can easily obtain a human's physiological data in concert with any other hardware or software nodes supporting \ac{ROS2}. 
%This is important, to show we are providing a metrically accurate equivalent. 
%To validate our package, we validate the wearable biosensor nodes by conducting comparison experiments with each manufacturer's program. 
To validate our package, we present plots of each sensor's published topics, in metric units. 
Finally, we present a potential application architecture using this biosensor package with a real-world robotic system.

\section{The Robot Operating System 2 (ROS 2)}
\label{sec:ros_introduce}
% What is ROS?
\ac{ROS}, a well-known open source middleware for robotics, assists in robotic system development by standardizing subsystem layouts and streamlining sensor communication through predefined communication rules, called standard \ac{ROS} messages \cite{quigley2009ros}. \ac{ROS} also provides various common tools for debugging and data visualization, decreasing system development time. 
Importantly, the ROS framework addresses the challenges and issues with data synchronization from various data formats (e.g., biosensors and image data), operating systems (e.g., Linux, Windows, and macOS), computer languages (e.g., C++, Python, and Java), and other frameworks (e.g., \ac{LSL} \cite{lsl_git}.
%Importantly, the \ac{ROS} framework addresses the challenges and issues with data synchronization from various sources (e.g., biosensors and image data), something that similar biosignal platforms, such as \ac{LSL} \cite{lsl_git}, lack. 
Since the introduction of \ac{ROS} in 2007, the robotics community has been vigorously developing the \ac{ROS} ecosystem, making it the preeminent framework in the field. However, \ac{ROS} was initially developed for single robot use, requiring excellent network quality among other limitations. As the robotics and \ac{HRI} fields expand, the need for real-time systems, for use in multi-robot and \ac{HRI} research, has arisen.

% Why Ver 2?
%sjm: There's no argument here for DDS as a transport vs ROSTCP, the transport used in ROS1
%Jo: I updated it by adding a reference
\ac{ROS2}, released in 2017, seeks to overcome many limitations of the initial \ac{ROS} architecture, such as official support of multi-robot designs \cite{ros2_advantages} and enabling real-time systems via \ac{DDS}, an open standard connectivity framework \cite{ros2_dds}. The major features in the \ac{ROS2} are as below: 

\begin{itemize}
    \item Support of real-time computing
    \item Support of multi-robot Systems
    \item Multiple nodes running in a process
    \item Distributed processing
    \item Flexible network environment
    %\item Support of the latest version of the computer languages
    \item Support of Windows 10
    \item Enhanced robot's security
\end{itemize}

%ROS1 TCP is reliable delivery as well. 
%It provides reliable publishing and subscribing \ac{ROS2} messages \cite{maruyama2016exploring} compared to TCPROS of the ROS 1 that is a transport layer in ROS 1 using the internet protocol suite (as known as TCP/IP). 
\ac{ROS2} increases accessibility to other disciplines through support for additional operating systems and programming languages, making cross discipline collaborations more realizable \cite{ros2_target_platform}. 
This cross-operating system support is particularly useful for biosensors, where a vendor may have only chosen to support a limited subset of modern operating systems and does not provide communications-level specifications to enable researchers to create an independent interface.
While the original version of \ac{ROS} is fully functional, the additional types of system architectures enabled through \ac{ROS2} provide flexibility to the system designer; typical components of \ac{HRI} research might include biosensors, processing nodes, actuators, interface nodes, and experimental monitoring, each of which may be running on separate computers and operating systems across continents or in the same room. 
%As a result of the new middleware design, \ac{ROS2} is not backwards compatible with \ac{ROS}, although limited cross-version communication is possible via ros1\_bridge.
Given the real-time multicomponent focus of \ac{ROS2} and the rise of biosensor usage in \ac{HRI} research,
%To be fair, no sensors have native ROS or ROS2 packages...
%and lack of \ac{ROS} compatibility,
a standardized \ac{ROS2} biosensor package is urgently needed.

\section{ROS2 Wearable Biosensor Package}
\label{sec:proposed_pacakge}

\begin{figure}[t] 
    \centering
    \includegraphics[width=1\columnwidth]{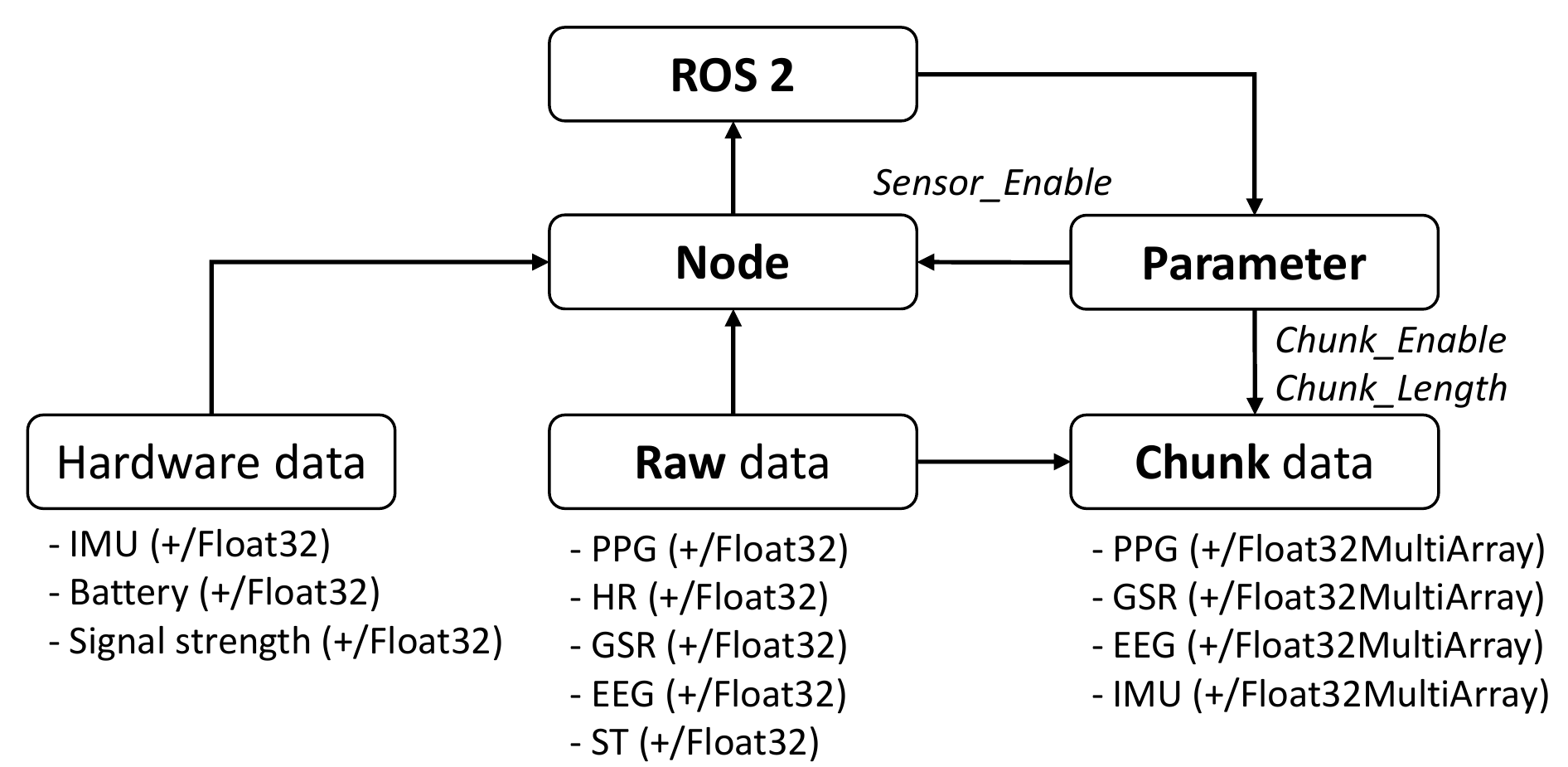}
    \small{\null\hfill  += \textit{standard\_msg}}
    \caption{Generalized structure of \ac{ROS2} nodes in the biosensor package.}
    \label{fig:biosensor_node_framework}
\end{figure}

Motivated by the multiple advantages of \ac{ROS2} (Section \ref{sec:ros_introduce}), we constructed a new wearable biosensor package for \ac{ROS2} focused on real-time functionality and ease of use, capable of impact in a broad range of research and development areas. The proposed \ac{ROS2} package (v0.0.1) currently supports six popular off-the-shelf wearable biosensors, each with their own node (Fig. \ref{img:supported_sensor_list}). We designed the biosensor package with the flexibility to expand to more wearable biosensors as desired. We will keep this package updated and add support for other wearable biosensors according to interest. The package (v0.0.1) is able at: \url{https://github.com/SMARTlab-Purdue/ros2-foxy-wearable-biosensors}

\subsection{Wearable Biosensor Framework}
Each package node follows a generalized structure, depicted in Fig. \ref{fig:biosensor_node_framework}. We categorized sensor data into three major types, using \ac{ROS2} standard messages and separated per node. Hardware data indicates current battery levels and Bluetooth signal strength. Nodes publish raw data in real-time, with sampling rates based on the individual biosensor hardware specifications. Chunk data, collected per node with predefined lengths, provide end-users with a framework for downstream processing (e.g., feature engineering and \ac{ML} applications). \ac{ROS2} parameters (\textit{Chunk\_Enable}, \textit{Chunk\_Length}, \textit{Sensor\_Enable}) control all three data types. \textit{Sensor\_Enable} and \textit{Chunk\_Enable} are Boolean data type (i.e., \textit{True} or \textit{False}), while \textit{Chunk\_Length} is an integer data type, adjusting data length per topic. Available topics depend on the individual biosensor hardware specifications. 

All package topic names follow the following format:

\begin{verbatim}
  /biosensors/<sensor_name>/<data_name>
\end{verbatim}

\noindent where the \textit{biosensor\_name} is the official name of the targeted biosensors (e.g., \textit{empatica\_e4}), and \textit{data\_name} is the biosignal type (e.g., \textit{PPG\_raw} and \textit{PPG\_chunk}).

\subsection{Supported Wearable Biosensors}
\begin{figure}[t]
    \centering
    \begin{subfigure}[b]{0.45\linewidth}
        \centering 
        \includegraphics[width=0.7\linewidth]{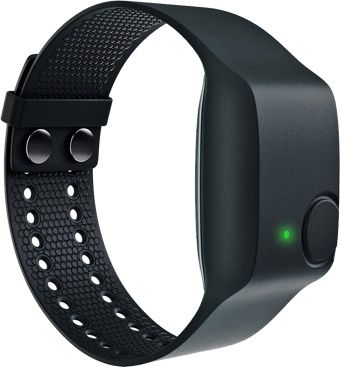} 
        \caption{}
        \label{img:empatica_e4}
    \end{subfigure}
    \begin{subfigure}[b]{0.45\linewidth}
        \centering 
        \includegraphics[width=1\linewidth]{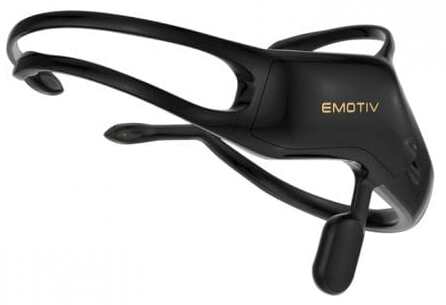} % need to remove the background.
        \caption{}
        \label{img:emotiv_insight}
    \end{subfigure}
    
    \begin{subfigure}[b]{0.45\linewidth}
        \centering 
        \includegraphics[width=0.8\linewidth]{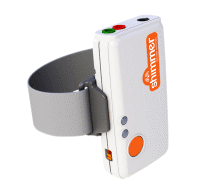} 
        \caption{}
        \label{img:shimeer_gsr}
    \end{subfigure}
    \begin{subfigure}[b]{0.45\linewidth}
        \centering 
        \includegraphics[width=1\linewidth]{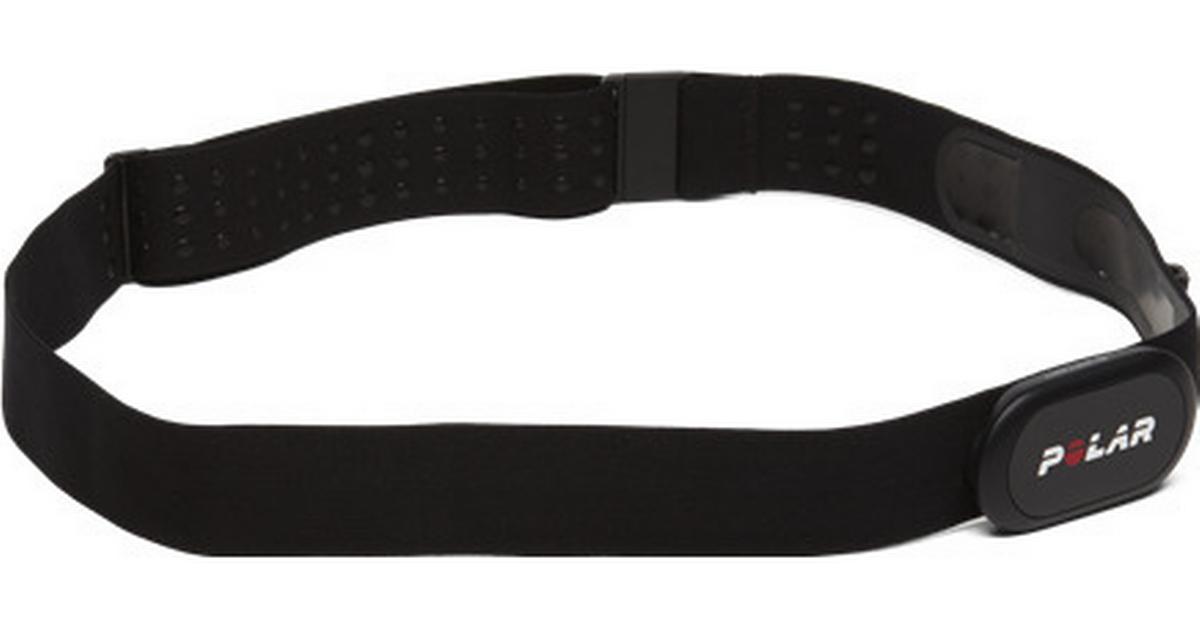} 
        \caption{}
        \label{img:polar_h10}
    \end{subfigure}
    
    \begin{subfigure}[b]{0.45\linewidth}
        \centering 
        \includegraphics[width=1\linewidth]{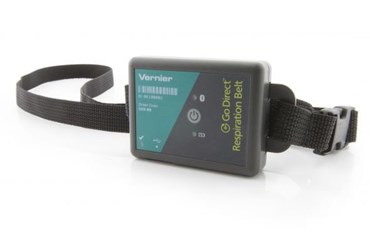} 
        \caption{}
        \label{img:vernier_respiration_belt}
    \end{subfigure}
    \begin{subfigure}[b]{0.45\linewidth}
        \centering 
        \includegraphics[width=1\linewidth]{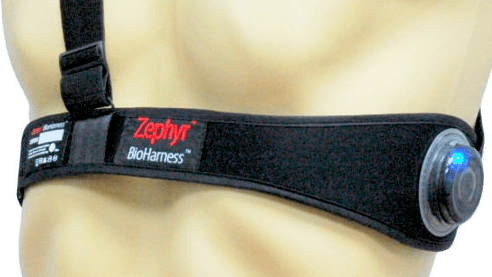} 
        \caption{}
        \label{img:zephyr_bioharness}
    \end{subfigure}
    \caption{A list of the biosensors that the proposed biosensor package currently supports; (a) Empatica E4-wristband \cite{empatica}, (b) EMOTIV Insight-5 Channel Mobile Brainwear \cite{emotiv}, (c) Shimmer3 GSR+ Unit \cite{shimmer}, (d) Polar H10-Heart rate monitor chest strap \cite{polarusa}, (e) Vernier-Go Direct Respiration Belt \cite{respriation_belt}, and (f) Zephyr BioHarness 3 \cite{zephyr}}
    \label{img:supported_sensor_list}
\end{figure}

\begin{figure*}[t] 
    \centering
    \begin{subfigure}{0.49\linewidth}
        \centering 
        \includegraphics[width=1\linewidth]{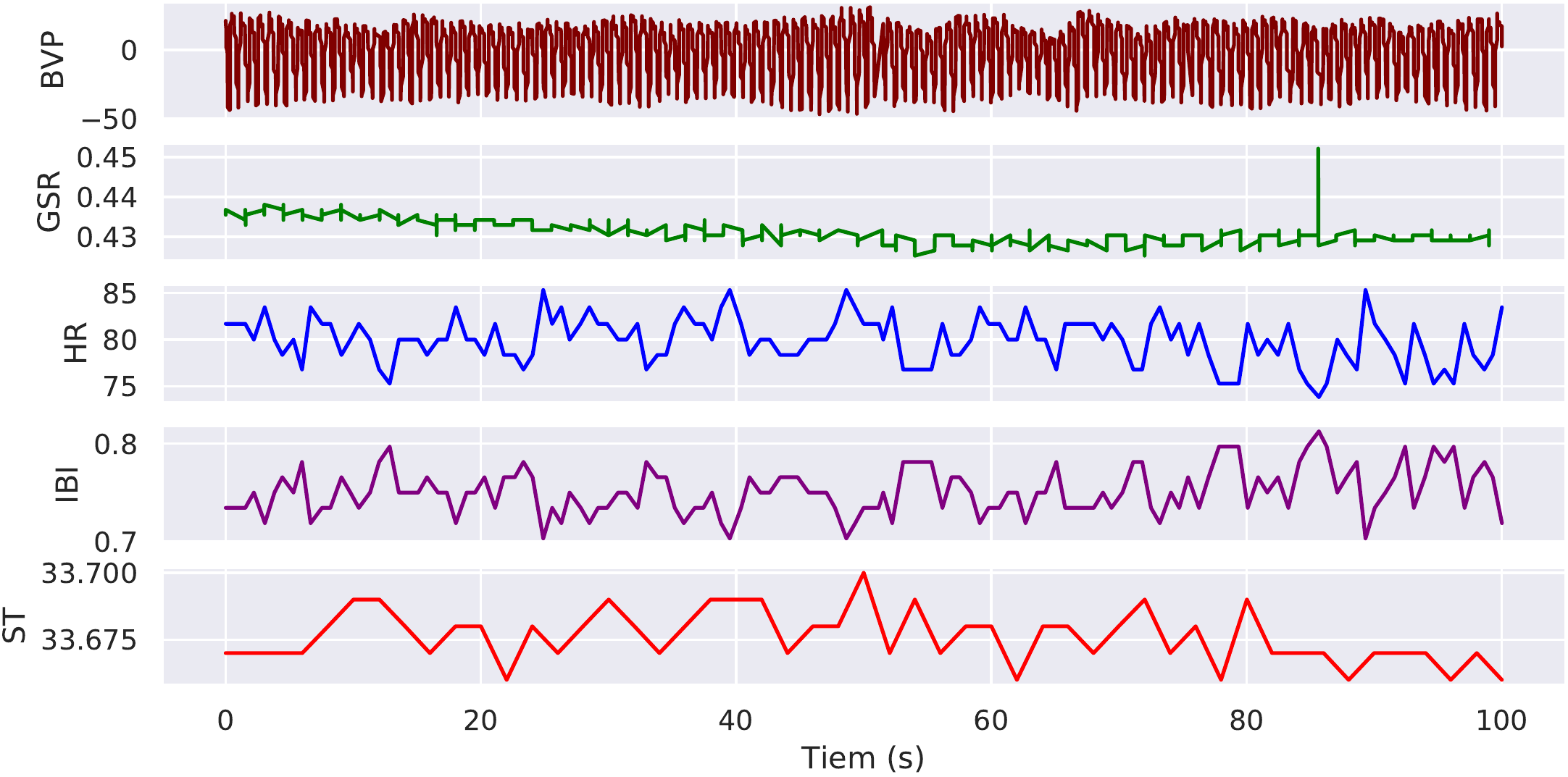} 
        \caption{}
        \label{img:empatica_data}
    \end{subfigure}
    \begin{subfigure}{0.49\linewidth}
        \centering 
        \includegraphics[width=1\linewidth]{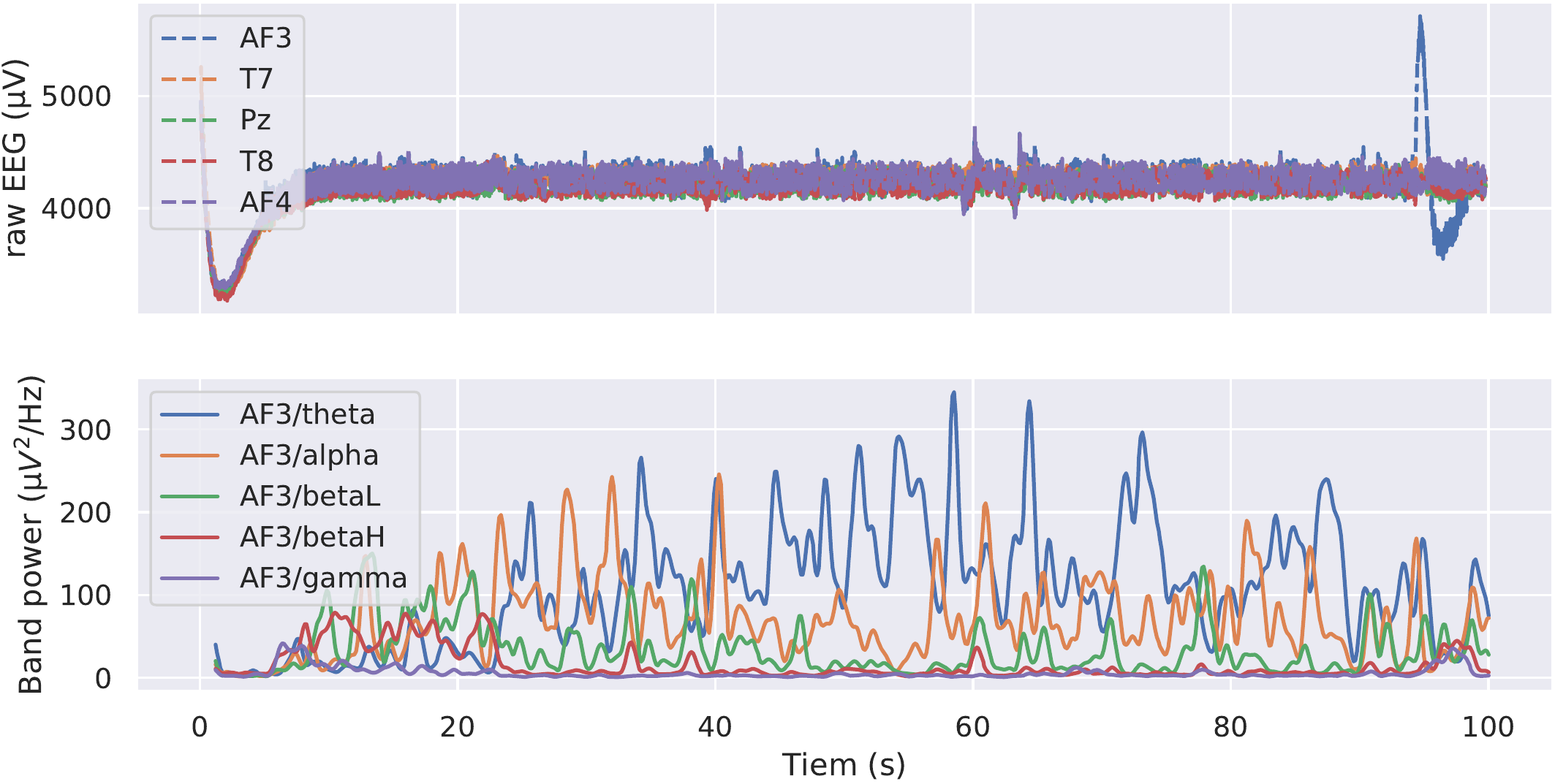} 
        \caption{}
        \label{img:emotive_data}
    \end{subfigure}
    % Now I am working on this sensor to gather data.
    \begin{subfigure}{0.49\linewidth}
        \centering 
        \includegraphics[width=1\linewidth]{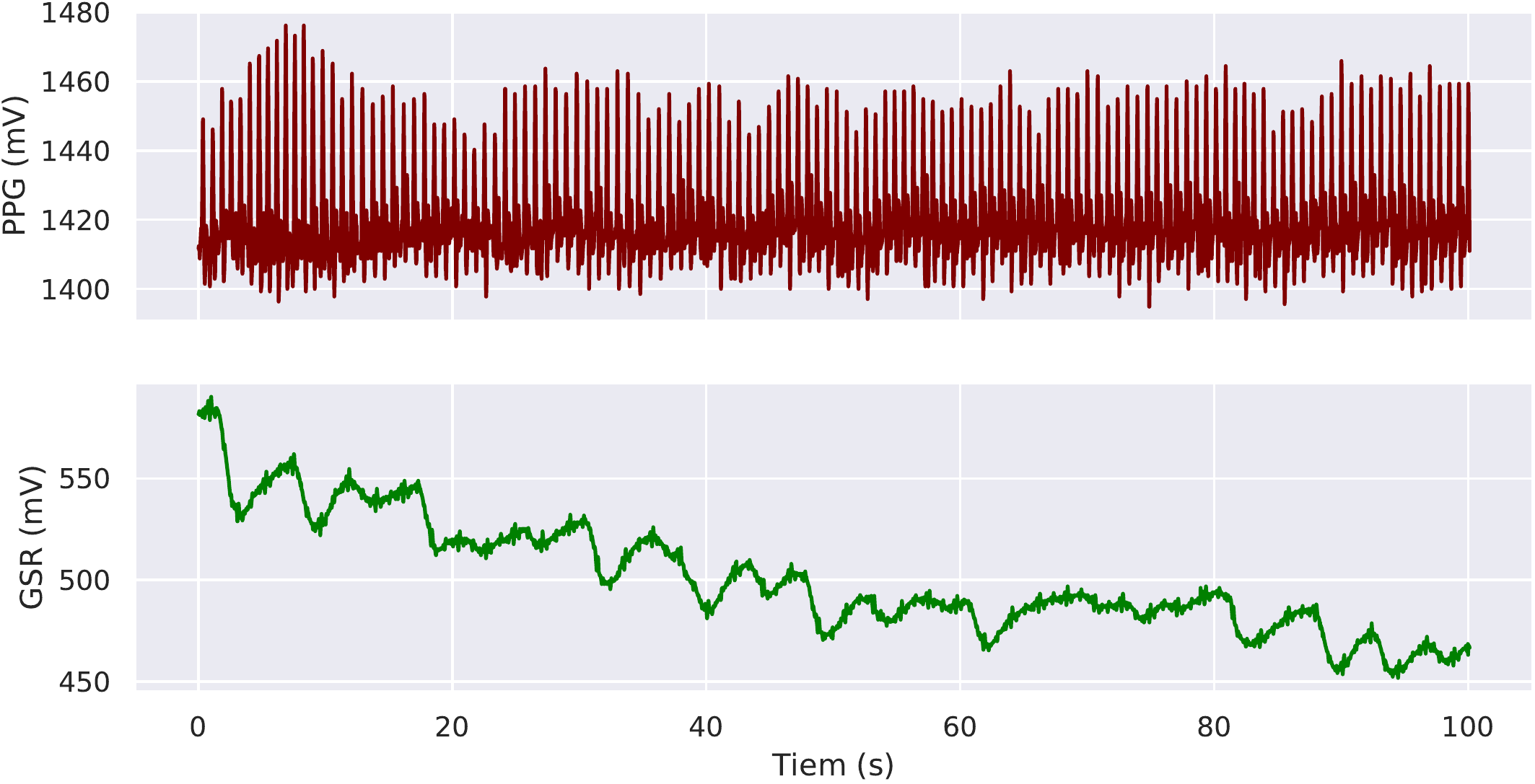} 
        \caption{}
        \label{img:shimmer3_data}
    \end{subfigure}
    \begin{subfigure}{0.49\linewidth}
        \centering 
        \includegraphics[width=1\linewidth]{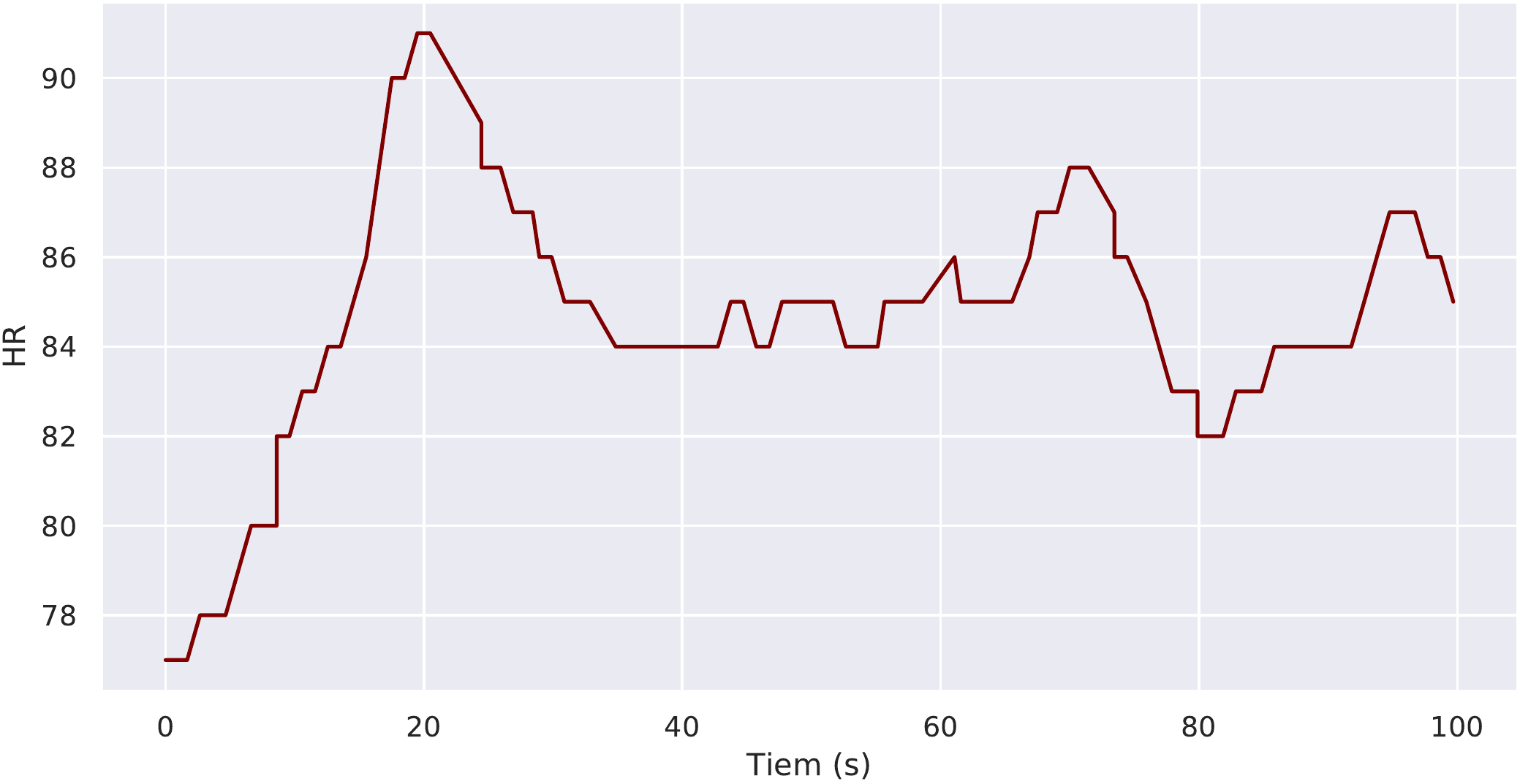}
        \caption{}
        \label{img:polar_data}
    \end{subfigure}
    
    \begin{subfigure}{0.49\linewidth}
        \centering 
        \includegraphics[width=1\linewidth]{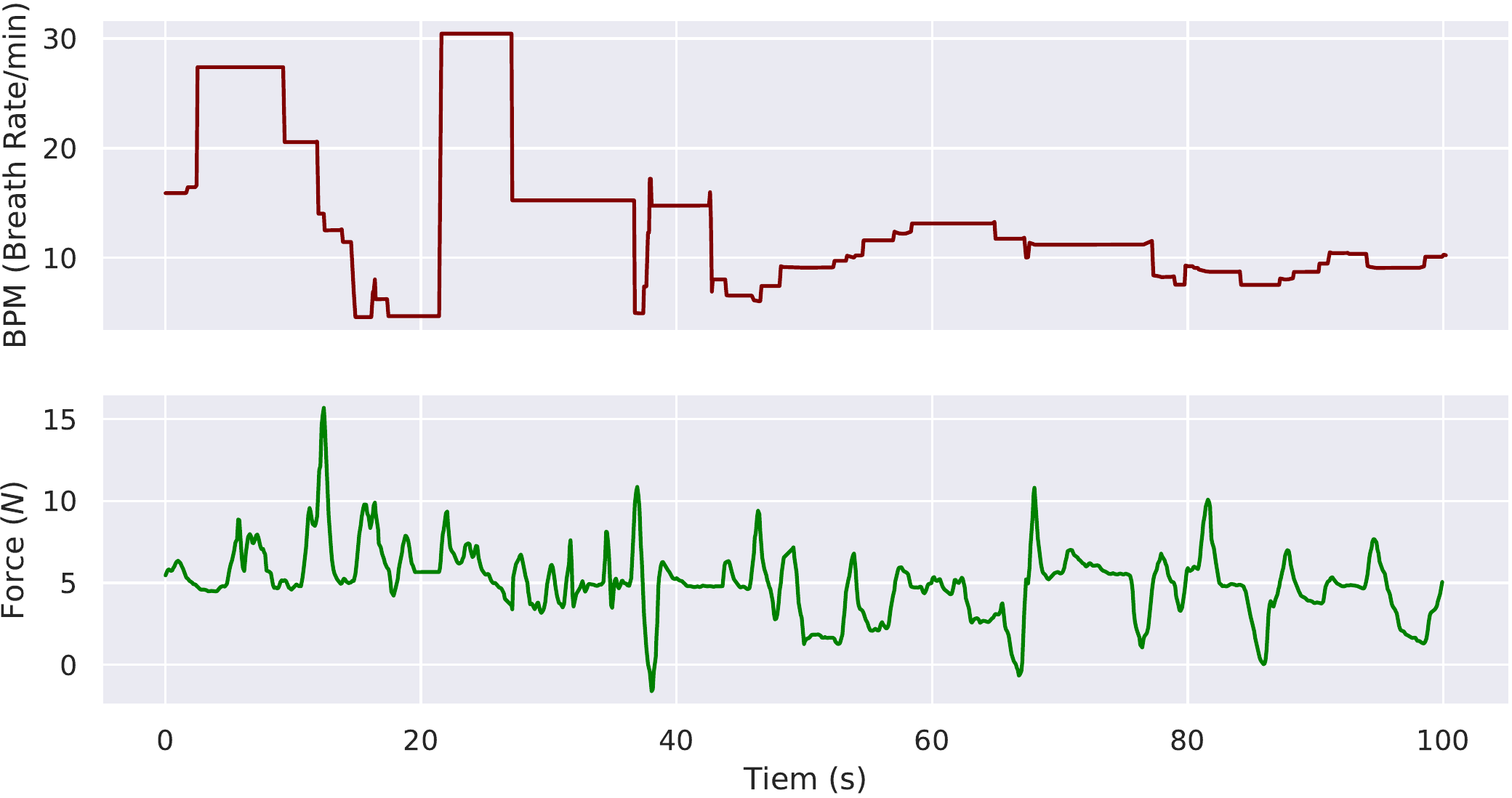} 
        \caption{}
        \label{img:vernier_data}
    \end{subfigure}
    \begin{subfigure}{0.49\linewidth}
        \centering 
        \includegraphics[width=1\linewidth]{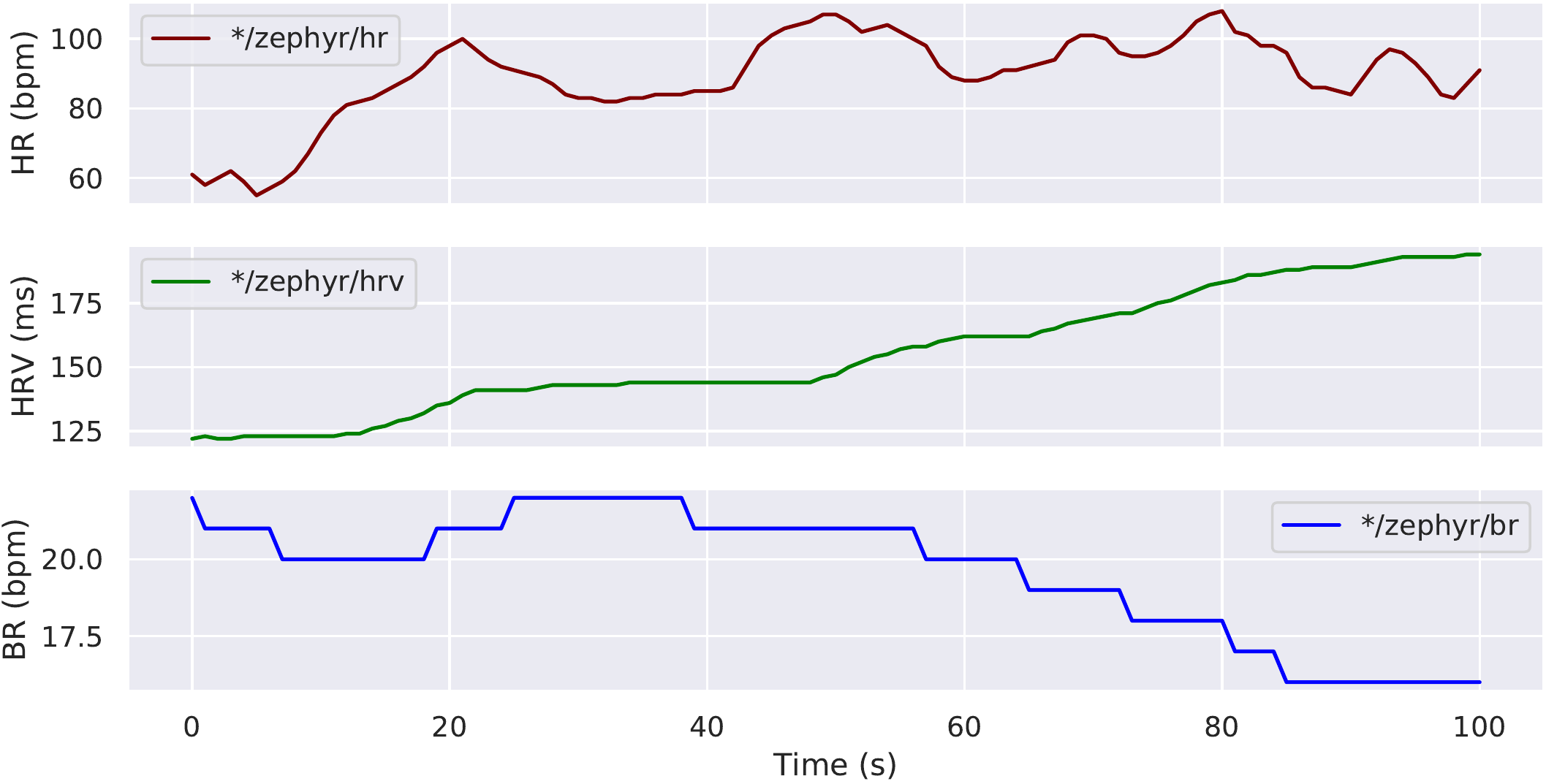} 
        \caption{}
        \label{img:zephyr_data}
    \end{subfigure}
    
    \caption{Collected biosignals from the published \ac{ROS2} topic data; (a) Empatica E4, (b) Emotive Insight, (c)Shimmer3-GSR Unit+, (d) Polar H10, (e) Vernier Respiration Belt, and (f) Zephyr Bioharness.}
    \label{img:wearable_data}
\end{figure*}

\subsubsection{Empatica E4 wristband} %(Purdue) for E4 streaming server for Windows​
%\resizebox{\columnwidth}{!}
The Empatica E4 is a wristband with an array of biosensors for monitoring biosignals: \ac{EDA} (or \ac{GSR}), \ac{BVP}, \ac{IBI}, \ac{HR}, and \ac{ST}, and behavioral monitoring: 3-axis accelerometer \cite{empatica} (Fig. \ref{img:empatica_e4}). 
However, there is a limitation to directly read biosensor data on the Linux environment since the vendor provides neither SDKs and libraries for Linux operating system nor technical interface documents. Thus, an additional Windows machine and Bluetooth dongle (e.g., Bluegiga Bluetooth Smart Dongle) are required in the current version of the biosensor package in order to stream biosensor data using \ac{LSL} as mentioned on the Empatica E4 website \cite{empatica_streaming_server}. The main Linux machine having the \ac{ROS2} system converts the \ac{LSL} data into \ac{ROS2} topics in real-time. The Empatica E4 node provides topics as depicted on Table \ref{tab:empatica_e4_topic_list}, and the Fig. \ref{img:empatica_data} is an example of the published data.

\begin{table}[t]
\caption{List of available topic names on the Empatica E4 Node.}
\label{tab:empatica_e4_topic_list}
\resizebox{\columnwidth}{!}{
    \begin{tabular}{llll}
    \hline
    \rowcolor[HTML]{C0C0C0}
    \textbf{Name} & \textbf{Type} & \multicolumn{1}{c}{\textbf{Topic name}} & \multicolumn{1}{c}{\textbf{Topic type}} \\ \hline
    Raw BVP             & raw       & \textit{*/bvp} & \textit{+/Float32} \\ \hline
    BVP Waveform       & chunk     & \textit{*/bvp\_chunk} & \textit{+/Float32MultiArray} \\ \hline
    Raw GSR             & raw       & \textit{*/gsr} & \textit{+/Float32} \\ \hline
    GSR Waveform      & chunk     & \textit{*/gsr\_chunk} & \textit{+/Float32MultiArray} \\ \hline
    Raw ST              & raw       & \textit{*/st} & \textit{+/Float32} \\ \hline
    ST Waveform            & chunk       & \textit{*/st\_chunk} & \textit{+/Float32MultiArray} \\ \hline
    Raw HR              & raw       & \textit{*/hr} & \textit{+/Float32} \\ \hline
    Raw IBI             & raw       & \textit{*/ibi} & \textit{+/Float32} \\ \hline
    Accelerometer   & hardware  & \textit{*/acc} & \textit{+/Float32MultiArray} \\ \hline
    Battery         & hardware  & \textit{*/bat} & \textit{+/Float32} \\ \hline
    Event button    & hardware  & \textit{*/tag} & \textit{+/Empty} \\ \hline
    \end{tabular}
}
    \null\hfill *= \textit{/biosensor/empatica\_e4} and += \textit{standard\_msg}.

\end{table}

\subsubsection{Emotiv Insight} %(Purdue) (license required)​
Emotiv Insight is a wearable headset capable of reading 8 channels \ac{EEG} signals (e.g., AF3, AF4, T7, T8, and Pz) (Fig. \ref{img:emotiv_insight}). The sampling rate of each channel is 128 samples per second with 14 bits resolution. It has an 9-axis inertial measurement unit (IMU) sensor to detect head motions \cite{emotiv}. Since it has a lightweight and user-friendly design, many affective researchers utilize it to measure the EEG signal from a human body \cite{jang2014development}. 

However, there is a limitation to stream and read raw \ac{EEG} data from the sensor without the Emotive Pro License, so the developers who want to use the Emotiv Insight device should have an Emotiv Pro license from Emotiv website \cite{emotiv_license}. Thus, we developed this Emotiv node with the Emotiv Pro license. For activating the Emotive Insight node, the main machine should be installed the EMOTIV App for Linux from \url{https://www.emotiv.com/my-account/downloads/}, then connected with the Emotiv Insight. 
The Emotiv Insight node provides topics as depicted in Table \ref{tab:emotiv_insight_topic_list}, where the band power includes the alpha, low beta, high beta, gamma, and theta bands, and the performance metrics are estimated by the Emotiv software that includes six metrics (defined by Emotiv): Stress (RUI), Engagement (ENG), interest (VAL), Excitement (EXC), Focus (FOC), and Relaxation (MED) \cite{emotiv_metrics}. Fig. \ref{img:emotive_data} is an example of the published data.

\begin{table}[t]
\caption{List of available topic names on the Emotiv Insight Node.}
\label{tab:emotiv_insight_topic_list}
\resizebox{\columnwidth}{!}{
    \begin{tabular}{llll}
    \hline
    \rowcolor[HTML]{C0C0C0}
    \textbf{Name} & \textbf{Type} & \multicolumn{1}{c}{\textbf{Topic name}} & \multicolumn{1}{c}{\textbf{Topic type}} \\ \hline
    EEG                 & raw       & \textit{*/eeg} & \textit{+/Float32MultiArray} \\ \hline
    EEG Waveform           & chunk     & \textit{*/eeg\_chunk} & \textit{+/Float32MultiArray} \\ \hline
    Band power          & raw       & \textit{*/pow} & \textit{+/Float32MultiArray} \\ \hline
    Band power Waveform   & chunk       & \textit{*/pow\_chunk} & \textit{+/Float32MultiArray} \\ \hline
    Performance metrics & raw       & \textit{*/met} & \textit{+/Float32MultiArray} \\ \hline
    Motion              & hardware  & \textit{*/mot} & \textit{+/Float32MultiArray} \\ \hline
    Device status      & hardware  & \textit{*/dev} & \textit{+/Float32MultiArray} \\ \hline
    \end{tabular}
}
    \null\hfill *= \textit{/biosensor/emotiv\_insight} and += \textit{standard\_msg}.

\end{table}

\subsubsection{Shimmer3-GSR Unit+} %(Purdue) 
The Shimmer3-GSR+ Unit is a wearable biosensor to measure \ac{GSR} and \ac{PPG} signals from the fingers or skins, converting to estimate HR \cite{shimmer} (Fig. \ref{img:shimeer_gsr}). The Shimmer3-GSR node provides topics as depicted on Table \ref{tab:shimmer3_topic_list}. Fig. \ref{img:shimmer3_data} is an example of the published data.

\begin{table}[t]
\caption{List of available topic names on the Shimmer3-GSR Node.}
\label{tab:shimmer3_topic_list}
\resizebox{\columnwidth}{!}{
    \begin{tabular}{llll}
    \hline
    \rowcolor[HTML]{C0C0C0}
    \textbf{Name} & \textbf{Type} & \multicolumn{1}{c}{\textbf{Topic name}} & \multicolumn{1}{c}{\textbf{Topic type}} \\ \hline
    GSR                 & raw       & \textit{*/gsr} & \textit{+/Float32} \\ \hline
    GSR Waveform          & chunk     & \textit{*/gsr\_chunk} & \textit{+/Float32MultiArray} \\ \hline
    PPG                 & raw       & \textit{*/ppg} & \textit{+/Float32} \\ \hline
    PPG Waveform      & chunk     & \textit{*/ppg\_chunk} & \textit{+/Float32MultiArray} \\ \hline
    \end{tabular}
}
    \null\hfill *= \textit{/biosensor/shimmer3\_gsr} and += \textit{standard\_msg}.
\end{table}

\subsubsection{Polar H10} %(Purdue) 
The Polar H10 is a wearable heart rate biosensor and attached on the chest (Fig. \ref{img:polar_h10}). It is mostly used for fitness objectives to read \ac{HR} with 1Hz sampling time \cite{polarusa}. The Polar H10 node provides topics as depicted in Table \ref{tab:polar_h10_topic_list}. Fig. \ref{img:polar_data} is an example of the published data.

\begin{table}[t]
\caption{List of available topic names on the Polar H10 Node.}
\label{tab:polar_h10_topic_list}
\resizebox{\columnwidth}{!}{
    \begin{tabular}{llll}
    \hline
    \rowcolor[HTML]{C0C0C0}
    \textbf{Name} & \textbf{Type} & \multicolumn{1}{c}{\textbf{Topic name}} & \multicolumn{1}{c}{\textbf{Topic type}} \\ \hline
    Raw HR                 & raw       & \textit{*/polar\_h10/hr} & \textit{+/Float32} \\ \hline
    \end{tabular}
}
    \null\hfill *= \textit{/biosensor} and += \textit{standard\_msg}.

\end{table}

\subsubsection{Vernier Respiration Belt} %(Purdue) 
The Vernier Respiration Belt is a wearable biosensor to measures human respiration rate from around the chest via Bluetooth (Fig. \ref{img:vernier_respiration_belt}). It is capable of measure from 0 to 50 N with 0.01 N resolution and breaths per minute (BPM) with 50 Hz sample rate \cite{respriation_belt}. The Vernier Respiration belt node provides topics as depicted on Table \ref{tab:vernier_topic_list}. Fig. \ref{img:vernier_data} is an example of the published data.

\begin{table}[t]
\caption{List of available topic names on the Vernier Respiration Belt Node.}
\label{tab:vernier_topic_list}
\resizebox{\columnwidth}{!}{
    \begin{tabular}{llll}
    \hline
    \rowcolor[HTML]{C0C0C0}
    \textbf{Name} & \textbf{Type} & \multicolumn{1}{c}{\textbf{Topic name}} & \multicolumn{1}{c}{\textbf{Topic type}} \\ \hline
    Raw BPM     & raw       & \textit{*/bpm} & \textit{+/Float32} \\ \hline
    BPM Waveform  & chunk     & \textit{*/bpm\_chunk} & \textit{+/Float32MultiArray} \\ \hline
    Raw Force   & raw     & \textit{*/force} & \textit{+/Float32} \\ \hline
    Force Waveform & chunk     & \textit{*/force\_chunk} & \textit{+/Float32MultiArray} \\ \hline

    \end{tabular}
    }
    \null\hfill *= \textit{/biosensor/veriner\_respiration\_belt} and += \textit{standard\_msg}.
\end{table}

\subsubsection{Zephyr Bioharness} %(UCSC)  
The Zephyr Bioharness is a chest strap sensor designed for dynamic movement activities \cite{zephyr} (Fig. \ref{img:zephyr_bioharness}). The Bioharness is capable of publishing output summary data (e.g., heart rate, acceleration) at 1 Hz. Raw ECG (FRQ = 256 Hz, Samples per msg = 63) and breathing (FRQ = 1.008 Hz, Samples per msg = 18) waveforms can be used in more advanced feature engineering. Currently, the Zephyr node provides topics as depicted in Table \ref{tab:zephyr_topic_list}. Fig. \ref{img:zephyr_data} is an example of the published data.

\begin{table}[t]
\caption{List of available topic names on the Zephyr Node.}
\label{tab:zephyr_topic_list}
\resizebox{\columnwidth}{!}{
    \begin{tabular}{llll}
    \hline
    \rowcolor[HTML]{C0C0C0}
    \textbf{Name} & \textbf{Type} & \multicolumn{1}{c}{\textbf{Topic name}} & \multicolumn{1}{c}{\textbf{Topic type}} \\ \hline
    Raw HR                 & raw       & \textit{*/hr} & \textit{+/uint8} \\ \hline
    Raw HRV                 & raw       & \textit{*/hrv} & \textit{+/uint16} \\ \hline
    ECG Waveform  & chunk     & \textit{*/ecg\_chunk} & \textit{+/Float32MultiArray} \\ \hline
    Raw BR                 & raw       & \textit{*/br} & \textit{+/Float32} \\ \hline
    BR Waveform   & chunk     & \textit{*/br\_chunk} & \textit{+/Float32MultiArray} \\ \hline
    \end{tabular}
}
    \null\hfill *= \textit{/biosensor/zephyr} and += \textit{standard\_msg}.

\end{table}

\iffalse
\subsection{ROS 2 Bag files for Data Analysis}
%%%%%% Need to update it.
The \ac{ROS2} system can save all topics including the biosensors, cameras, as well as the subjective questionnaires to a single ROS2bag file with synchronized time. Thus, the ROS2bag has more benefits than a traditional \ac{CSV} file regarding of collecting and analyzing the data set. Since the ROS can ensure to synchronize the recording all topic data, it is available to easily and directly analyze the data set by replaying both using a single ROS2bag file. Also, users can validate the developing algorithm and programs by launching the ROS2bag file \cite{jo2020rosbag}.
\fi

\section{Potential Application}
\begin{figure}[t] 
  \centering
  \includegraphics[width=1\columnwidth]{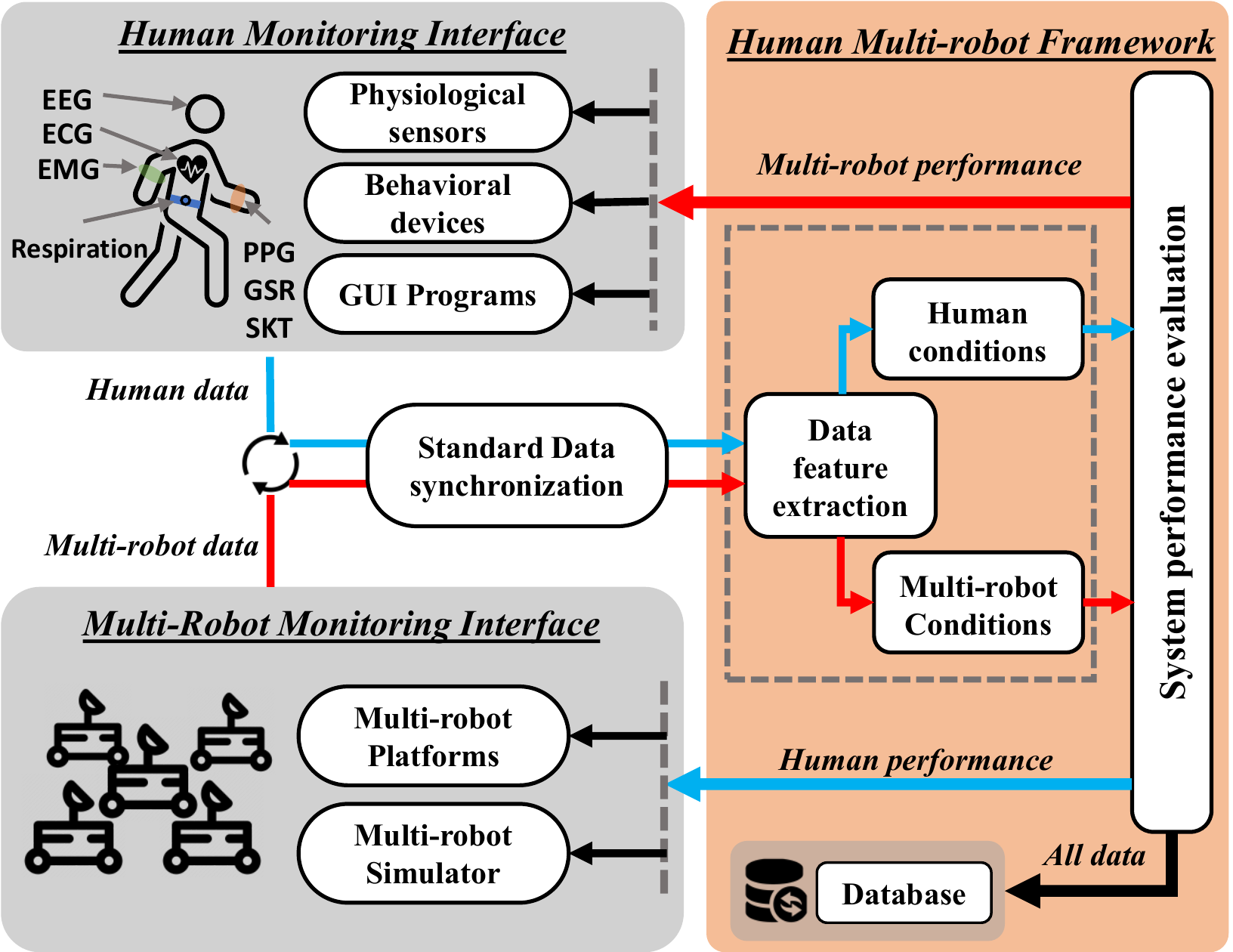}
   \caption{A potential application for \ac{HMRI} systems \cite{jo2020ros}.}
   % \vspace{-0.5em}
   \label{fig:potential_application}
\end{figure}

A standardized biosensor package for \ac{ROS2} will play a vital role in \ac{HMRI} research. One of the potential applications is the configuration of a \ac{HMRI} processing framework as illustrated in Fig. \ref{fig:potential_application}. The system could monitor multiple human and robot states \cite{jo2020rosbag, jo2020ros} and adjust individual task allocations as needed \cite{mina2020adaptive, mcguire2019everybody}. A system designed for such use could consist of four discrete elements: 1) human and robot condition monitoring, 2) feature extraction from raw data, 3) data storage using ROS2bag, and 4) evaluation, which would include action adjustments and data visualization. Biosensor data would provide human physiological and behavioral observations, while various operation parameters (e.g. battery level, encoder positions, and internal temperature) would provide insight into the real-time robot states. New or existing libraries (for example, pyphysio \cite{bizzego2019pyphysio}, NeuroKit2 \cite{Makowski2021neurokit}, and BioSPPy \cite{carreiras2015biosppy}) would extract physiologically relevant features. A \ac{ML} model could use the collected data as input and produce an adjustment to the current system. By using \ac{ROS2}'s recording system, all subscribed/published data can be stored in a single ROS2bag file with synchronized timestamps between measurements, simplifying data analysis and management \cite{jo2020rosbag}.
%It is helpful and useful to analyze the dataset by replaying both using a single ROS2bag file \cite{jo2020rosbag}. 
%An overview of a potential \ac{HMRI} framework is 

\section{Conclusion and Future work}
\label{sec:conclusion_futureworks}
In this paper, we introduced a new package to integrate wearable biosensors into the \ac{ROS2} ecosystem that can facilitate to easily measure and utilize the predicted human's affective states (such as emotional and cognitive states) on the \ac{HRI} and \ac{HMRI} applications. Each node of the package communicates using \ac{ROS2} in real-time, a distinct advantage of \ac{ROS2}.  The proposed package (v0.0.1) currently contains six off-the-shelf wearable biosensors. A generalized node and topic layout govern each node's structure. The \ac{ROS2} parameters control published topic information, such as biosensor data channel activation and chunk data length. To demonstrate the performance of the biosensor package, we presented representative data set examples collected from each sensor using \ac{ROS2}.

%In this paper, we introduced a new package to integrate wearable biosensors into the \ac{ROS2} ecosystem. Each node of the package communicates using \ac{ROS2} in real-time, a distinct advantage of \ac{ROS2}. Also, the ROS2 help other devices to connect with robotic platforms. The proposed package (v0.0.1) currently contains six different biosensors. A generalized node and topic layout govern each node's structure. \ac{ROS2} parameters control published topic information, such as biosensor data channel activation and chunk data length. To demonstrate the performance of the biosensor package, we presented representative data set examples collected from each sensor using \ac{ROS2}.

%%% Future works
In the future, we intend to integrate more wearable biosensors and topics in accordance with community or industry interest and welcome outside contribution as the \ac{ROS2} ecosystem expands. We plan to keep this package up-to-date and create a Docker image for an even easier environment initialization. Thus, as robotics research becomes more interdisciplinary and complex, we will position this package to be a fundamental resource for the \ac{HRI} community by providing standardized tools and a minimal barrier to entry. We look forward to the future innovations this package will foster. %%%%%%%%%%%%%%%%%%%%%%%%%%%%%%%%%%%%%%%%%%%%%%%%%%%%%%%%%%%%%%%%%%%%%%%%%%%%%%%%

\section*{Acknowledgements}
This material is based upon work supported by the National Science Foundation under Grant No. IIS-1846221 and by DARPA under grant HR0011-18-2-0043. Any opinions, findings, and conclusions or recommendations expressed in this material are those of the author(s) and do not necessarily reflect the views of the National Science Foundation or DARPA.

\bibliography{main}
\bibliographystyle{IEEEtran}

\end{document}